\documentclass[10pt,twocolumn,letterpaper]{article}

\usepackage{wacv}              

\usepackage{multirow}
\usepackage[table]{xcolor}

\definecolor{wacvblue}{rgb}{0.21,0.49,0.74}
\usepackage[pagebackref,breaklinks,colorlinks,allcolors=wacvblue]{hyperref}
\usepackage[accsupp]{axessibility} 
\def\wacvPaperID{750} 
\def\confName{WACV}
\def\confYear{2027}

\title{Sidecar: Training-Free Semantic Reuse for \\ Character-Consistent Free-form Visual Storytelling}

\author{Sibo Dong \qquad  Sarah Adel Bargal \\
Department of Computer Science, Georgetown University, Washington, D.C., USA\\
{\tt\small \{sd1242, sarah.bargal\}@georgetown.edu}
}

\begin{document}
\maketitle

\begin{abstract}
Visual storytelling requires generating images that follow a narrative while preserving consistent character identities across frames. In free-form story generation, a character is fully described only when first introduced and is later referred to by a type-level mention or pronoun. Although this setting better reflects natural storytelling, later prompts may omit important identity-related semantics, making character consistency more difficult to maintain.
We propose \textbf{Sidecar}, a plug-and-play semantic augmentation module that preserves entity-level information from the initial description and injects the missing semantics into later prompt embeddings. Sidecar requires no additional training and does not modify the architecture of the base diffusion model. Experiments on FreeStoryBench show that Sidecar consistently improves prompt-image alignment and character consistency across multiple SDXL- and FLUX-based baselines, with negligible computational overhead.
\end{abstract}

\section{Introduction}
Visual storytelling aims to generate a sequence of images from a sequence of narrative prompts. Each image should accurately reflect its corresponding prompt, while the complete sequence should preserve consistent characters, objects, and visual appearance across frames. Compared with single-image text-to-image (T2I) generation, visual storytelling requires the model to satisfy both prompt-image alignment and cross-frame consistency. However, pretrained T2I models \cite{sd,sdxl,sd3,flux} typically generate each image independently and lack an explicit mechanism for preserving the cross-frame consistency.

Existing visual storytelling methods improve cross-frame consistency by incorporating information from previous prompts or generated images~\cite{arldm,makeastory,acmvsg,storygen,storygptv,vista}. Several approaches introduce additional modules trained on story-level data to capture dependencies across frames~\cite{causalstory,seedstory,storyimager,thechosenone,cogcartoon,talecrafter}. Although effective, these methods require task-specific training and additional computational resources. Training-free methods instead modify the inference process of pretrained T2I models~\cite{consistory,storydiffusion,characonsist,onepromptonestory,freestory} to preserve character consistency by reusing attention features or latent representations from previous generations.

\begin{figure*}
    \centering
    \includegraphics[width=0.9\linewidth]{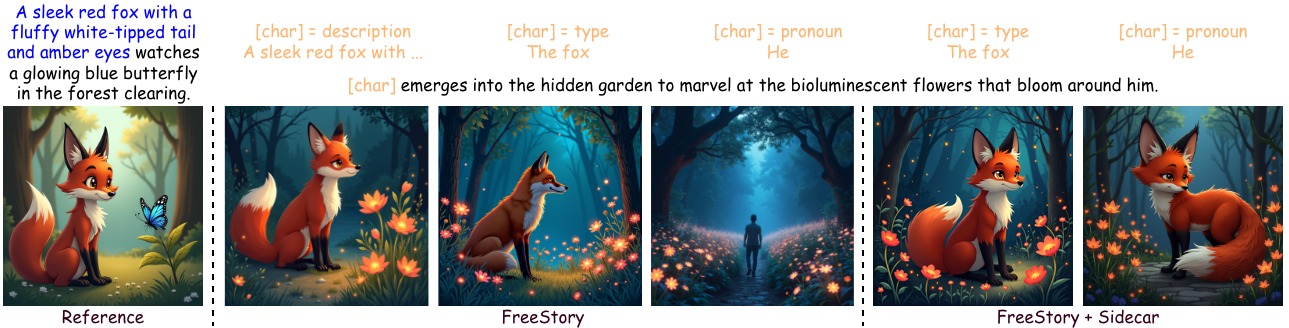}
    \caption{\textbf{Identity drift under different character references in free-form story generation.} Starting from the same initial character, we generate the next frame using a state-of-the-art free-form visual storytelling method, FreeStory~\cite{freestory}, while keeping generation settings unchanged. Replacing the full description with a type or pronoun progressively removes explicit identity-related information from the prompt, leading to inconsistency. By incorporating our Sidecar method, improved character consistency is achieved under both settings.
}
    \label{fig:semantic_degradation}
\end{figure*}

Most training-free methods assume that the complete character description is repeated in every prompt. Repeated descriptions provide the text encoder with similar character semantics, which simplifies identity preservation. However, this prompt format differs from natural storytelling, where a character is generally described upon first appearance and later referred to by a name, a type, or a pronoun.

FreeStory~\cite{freestory} addresses this by introducing the free-form story generation setting. Under this formulation, the complete character description is included only in the first prompt, while subsequent prompts use shorter references such as a character type or a pronoun. For example, after introducing \textit{``a sleek red fox with a fluffy white-tipped tail and amber eyes,''} later prompts may refer to the same character as \textit{``the fox''} or \textit{``he.''} This formulation more closely follows natural narrative structure and avoids repeatedly inserting the same description into every prompt. It also improves the readability and semantic continuity of the narrative. Moreover, shorter references reduce prompt length, which is particularly important for long or multi-character stories, where repeated descriptions may occupy a substantial portion of the text encoder's maximum length.

Although FreeStory improves character consistency under free-form prompts, later references can still weaken the text condition. As shown in Figure~\ref{fig:semantic_degradation}, replacing a full character description with a type-level mention or pronoun removes explicit identity attributes, leaving later prompt embeddings semantically incomplete. Visual feature reuse~\cite{consistory,storydiffusion,freestory,onepromptonestory,characonsist} can transfer information from previous frames, but it does not directly restore the missing identity semantics in the current text condition.

To address this, we propose \textbf{Sidecar}, a plug-and-play semantic augmentation module for free-form visual storytelling. Sidecar preserves entity-level semantic information from the initial character description and injects the missing character semantics into the text representations of later prompts. Rather than replacing the original text embedding or appending the full description to every prompt, Sidecar provides an auxiliary semantic representation alongside the original text-encoding process. The original prompt specifies the action and scene, while Sidecar supplements the identity information omitted by the character reference. Sidecar requires no additional training and does not modify the architecture of the base diffusion model. It can therefore be integrated into different pretrained generation pipelines and combined with existing storytelling methods.

We evaluate Sidecar on FreeStoryBench \cite{freestory} using both SDXL- and FLUX-based models. Sidecar consistently improves vanilla  T2I backbones and existing storytelling methods, including StoryDiffusion~\cite{storydiffusion}, ConsiStory~\cite{consistory}, and FreeStory~\cite{freestory}. The results demonstrate improvements in both prompt-image alignment and character consistency across story frames. Sidecar also introduces negligible additional GPU memory usage and inference time.

In summary, Sidecar provides a plug-and-play mechanism for preserving entity-level semantics under free-form story prompts. It injects the character information omitted by later mentions into the text-conditioning process without changing the architecture of the base diffusion model. Experiments on FreeStoryBench show consistent improvements across multiple SDXL- and FLUX-based baselines, with negligible additional computational overhead.

\begin{figure*}[t]
    \centering
    \includegraphics[width=0.95\linewidth]{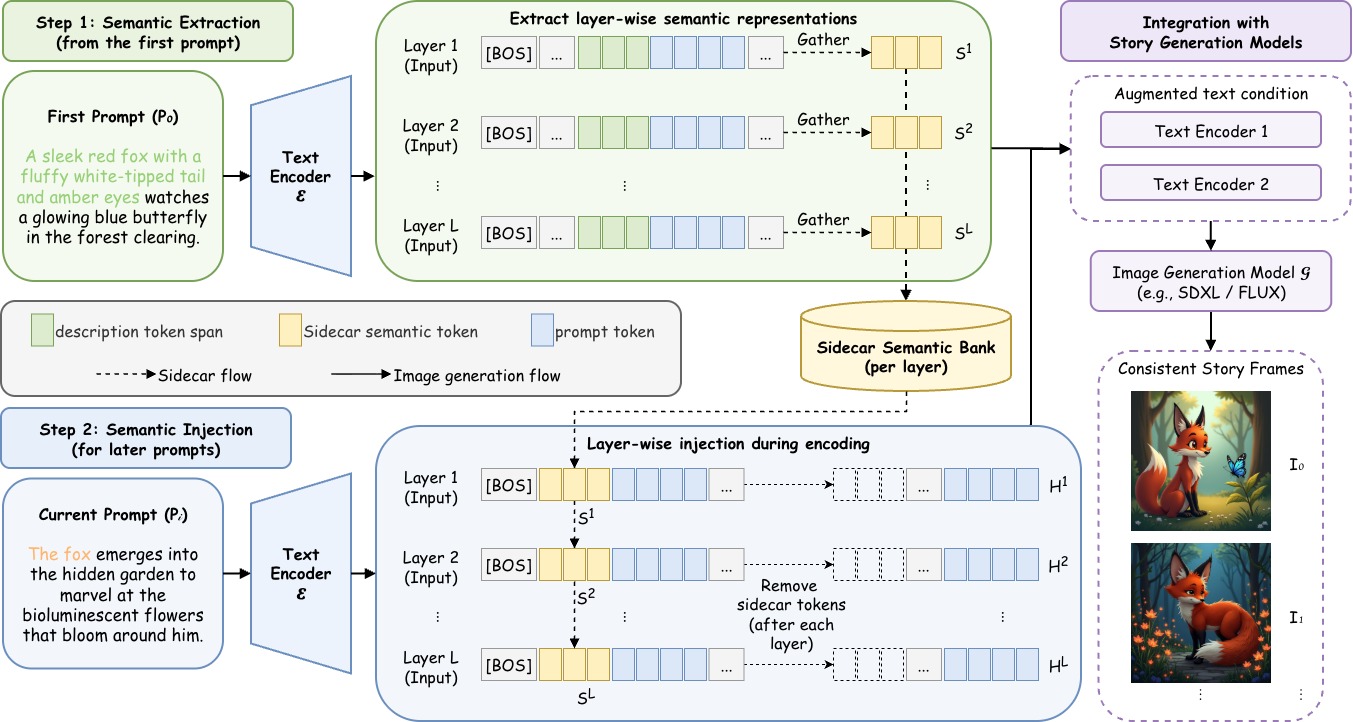}
    \caption{\textbf{Overview of Sidecar.} Given the first story prompt, Sidecar extracts the hidden states corresponding to the complete character description at each text-encoder layer and stores them as layer-wise semantic representations. When encoding a later prompt, the corresponding representations are temporarily inserted after the begin-of-sentence token and interact with the current prompt tokens through self-attention. The temporary tokens are removed after each layer, preserving the original output length and interface of the text encoder. The resulting augmented text condition is then provided to the original image-generation pipeline, while any existing visual consistency mechanism remains unchanged.}
    \label{fig:overview}
\end{figure*}

\section{Related Work}

\subsection{Visual Storytelling}

Visual storytelling aims to generate a sequence of images that follows a textual narrative while preserving coherent characters, scenes, and visual appearance across frames. Early methods learn story-level dependencies by conditioning each generation on previous prompts or images \cite{arldm,makeastory,acmvsg, seedstory}, or by introducing additional modules to aggregate multimodal history~\cite{talecrafter,cogcartoon,causalstory,thechosenone,storygen,storygptv,vista}. These approaches improve visual consistency through task-specific training on story datasets. However, they require additional trainable components, story-level supervision, or modifications to the original generation architecture. These limitations motivate training-free methods that adapt pretrained T2I models directly during inference.

\subsection{Training-Free Consistent Story Generation}

A major line of training-free research preserves subject identity by reusing attention or intermediate visual features across generated images. ConsiStory~\cite{consistory} introduces subject-driven shared attention and correspondence-based feature injection to transfer subject information between images. StoryDiffusion~\cite{storydiffusion} proposes Consistent Self-Attention, which establishes feature interactions across a sequence of generations. CharaConsist~\cite{characonsist} further improves fine-grained character consistency through point-tracking attention, adaptive token merging, and decoupled control of foreground and background features. 

Another line of work improves consistency by modifying, expanding, or reorganizing the textual conditions used for story generation. OnePromptOneStory~\cite{onepromptonestory} concatenates all story prompts into a single input and exploits the contextual consistency of the text encoder to preserve character identity.
Infinite-Story~\cite{infinitestory} introduces Identity Prompt Replacement to align identity attributes across different prompts, together with attention guidance for appearance and style consistency. DreamStory~\cite{dreamstory} uses an LLM to construct detailed descriptions for the subjects and scenes. It then generates subject portraits and uses them together with the textual descriptions as multimodal references. 
These methods demonstrate the importance of textual and multimodal conditions to improve consistency. However, they commonly concatenate, rewrite, replace, or expand the original prompts. 
In contrast, Sidecar retains the original narrative prompts and supplements their text representations with entity-level semantics preserved from the initial description.

Most training-free consistency methods assume that a complete character description is repeated across prompts \cite{onepromptonestory} or can be reconstructed through prompt rewriting \cite{dreamstory}. FreeStory~\cite{freestory} introduces a free-form story setting in which the full character description is provided only in the first prompt, while later prompts refer to the same character through a type-level mention or pronoun. This setting more closely follows natural narrative structure and avoids repeatedly inserting detailed descriptions into every prompt. FreeStory maintains character consistency by reusing attention features from previous generations. Nevertheless, the text representations of later prompts remain semantically under-specified because type and pronouns omit the detailed identity information contained in the initial prompt. Sidecar addresses this complementary text-side limitation by preserving the entity semantics from the first prompt and injecting them into later text representations without rewriting the prompts or modifying the base diffusion model.

\section{Method}

\subsection{Problem Formulation}

Given a story consisting of $N$ textual prompts, we denote the prompt sequence as
\begin{equation}
    \mathcal{P} = \{P_0, P_1, \ldots, P_{N-1}\}.
\end{equation}
Under the free-form setting, the character is introduced with a detailed description in the first prompt, while subsequent prompts refer to the same character using a shorter type-level mention or pronoun. Let $d$ denote the complete character description and $m_i$ denote its reference in prompt $P_i$. Although $d$ and $m_i$ refer to the same entity, they contain substantially different semantic information.

A pretrained text encoder $\mathcal{E}$ maps each prompt to a sequence of contextualized token representations:
\begin{equation}
    \mathbf{H}_i
    =
    \mathcal{E}(P_i)
    \in
    \mathbb{R}^{L \times D},
\end{equation}
where $L$ is the token sequence length and $D$ is the hidden dimension. When $P_i$ contains only a type or pronoun, its representation may not retain the appearance attributes introduced by $d$. Existing storytelling methods can transfer visual information from previous generations, but the current text condition remains semantically under-specified.

We address this by augmenting the text-encoding process with the hidden representations of the initial character description. The augmented text condition is denoted as
\begin{equation}
    \widetilde{\mathbf{H}}_i
    =
    \mathcal{E}_{\mathrm{Sidecar}}(P_i,d).
\end{equation}
Sidecar does not rewrite the current prompt or permanently increase the length of its output representation. Instead, it introduces temporary semantic tokens within the text encoder, allowing the current prompt to retrieve the identity information omitted by the shorter character mentions.

\subsection{Sidecar Semantic Augmentation}

Figure~\ref{fig:overview} presents an overview of Sidecar. The method consists of two stages: semantic extraction and semantic injection. During semantic extraction, Sidecar identifies the detailed character description and extracts its layer-wise hidden representations. During semantic injection, these representations are inserted as temporary semantic tokens when encoding later prompts.

\paragraph{Semantic Extraction.}

We first identify the token span corresponding to the complete character description $d$. Let
\begin{equation}
    \mathcal{M}
    =
    \{(s_1,e_1),\ldots,(s_K,e_K)\}
\end{equation}
denote the set of matched token spans, where each pair $(s_k,e_k)$ specifies the start and end indices of a description phrase. 
Consider a Transformer text encoder with $L$ layers. Let
\begin{equation}
    \mathbf{H}^{\ell-1}_{\mathrm{ref}}
    \in
    \mathbb{R}^{L \times D}
\end{equation}
denote the input hidden states of layer $\ell$ when encoding the initial character description. At each layer, Sidecar extracts the hidden states corresponding to the identified character:
\begin{equation}
    \mathbf{S}^{\ell}
    =
    \operatorname{Gather}
    \left(
        \mathbf{H}^{\ell-1}_{\mathrm{ref}},
        \mathcal{M}
    \right)
    \in
    \mathbb{R}^{M \times D},
    \label{eq:sidecar_extract}
\end{equation}
where $M$ is the total number of tokens contained in the matched spans.

The extracted representation $\mathbf{S}^{\ell}$ preserves the token-level semantics of the full character description at layer $\ell$. Sidecar stores these representations separately for each encoder layer, retaining the contextualized character information formed at different representation depths.

\paragraph{Semantic Injection.}

When encoding a later prompt $P_i$, Sidecar injects the extracted description representations into each Transformer layer. Let
\begin{equation}
    \mathbf{H}^{\ell-1}_{i}
    =
    \left[
        \mathbf{h}^{\ell-1}_{i,\mathrm{BOS}};
        \mathbf{h}^{\ell-1}_{i,1};
        \ldots;
        \mathbf{h}^{\ell-1}_{i,L-1}
    \right]
\end{equation}
denote the input hidden states of layer $\ell$, where
$\mathbf{h}^{\ell-1}_{i,\mathrm{BOS}}$ is the begin-of-sentence representation. Sidecar inserts $\mathbf{S}^{\ell}$ immediately after the begin-of-sentence token:
\begin{equation}
    \mathbf{X}^{\ell}_{i}
    =
    \left[
        \mathbf{h}^{\ell-1}_{i,\mathrm{BOS}};
        \mathbf{S}^{\ell};
        \mathbf{h}^{\ell-1}_{i,1};
        \ldots;
        \mathbf{h}^{\ell-1}_{i,L-1}
    \right].
    \label{eq:sidecar_injection}
\end{equation}

The augmented sequence is then processed by the original Transformer layer:
\begin{align}
    \overline{\mathbf{X}}^{\ell}_{i}
    &=
    \mathbf{X}^{\ell}_{i}
    +
    \operatorname{MSA}^{\ell}
    \left(
        \operatorname{LN}^{\ell}_1
        \left(
            \mathbf{X}^{\ell}_{i}
        \right)
    \right), \\
    \widehat{\mathbf{X}}^{\ell}_{i}
    &=
    \overline{\mathbf{X}}^{\ell}_{i}
    +
    \operatorname{MLP}^{\ell}
    \left(
        \operatorname{LN}^{\ell}_2
        \left(
            \overline{\mathbf{X}}^{\ell}_{i}
        \right)
    \right).
    \label{eq:sidecar_transformer}
\end{align}

The semantic tokens are visible to the tokens of the current prompt during self-attention. The prompt representation can therefore retrieve identity-related information that is absent from a type-level mention or pronoun. Unlike direct embedding replacement, this operation allows the text encoder to contextualize the initial character semantics together with the current action, scene, and other content.

After the Transformer layer is evaluated, the temporary semantic-token outputs are removed:
\begin{equation}
    \mathbf{H}^{\ell}_{i}
    =
    \operatorname{RemoveSidecar}
    \left(
        \widehat{\mathbf{X}}^{\ell}_{i}
    \right).
    \label{eq:sidecar_remove}
\end{equation}
The resulting representation has the same sequence length and hidden dimension as the original prompt representation. It can therefore be passed to the next encoder layer without changing the external interface of the text encoder.

This process is repeated at every text-encoder layer using the semantic representation extracted from the corresponding layer. In particular, $\mathbf{S}^{\ell}$ is inserted only during the computation of layer $\ell$ and is removed before the next layer. The semantic tokens are therefore temporary and do not accumulate across the encoder.

After the final layer, Sidecar produces the augmented prompt representation
\begin{equation}
    \widetilde{\mathbf{H}}_i
    =
    \mathcal{E}_{\mathrm{Sidecar}}
    \left(
        P_i,
        \{\mathbf{S}^{\ell}\}_{\ell=1}^{L_{\mathcal{E}}}
    \right).
\end{equation}
The original narrative prompt remains unchanged. Sidecar does not replace mentions such as ``the man'' or ``he'' with the full description. Instead, the current prompt specifies the new action and scene, while the temporary semantic tokens provide the identity information omitted by the reference.

\subsection{Encoder-Specific Integration}

Sidecar operates within the text-conditioning stage and does not modify the image-generation network. Let $\mathcal{G}$ denote a pretrained image generator and let $\mathcal{V}_{<i}$ represent any history visual information used by an existing visual storytelling method \cite{storydiffusion,consistory,onepromptonestory,freestory}. The $i$-th story frame is generated as
\begin{equation}
    I_i
    =
    \mathcal{G}
    \left(
        \widetilde{\mathbf{H}}_i,
        \mathcal{V}_{<i}
    \right),
    \label{eq:sidecar_generation}
\end{equation}
where $\mathcal{V}_{<i}$ is omitted for a vanilla text-to-image model. When Sidecar is combined with existing methods, their original consistency mechanisms remain unchanged. Sidecar operates as a complementary text-side module that augments the semantic condition, while the underlying method continues to propagate visual features across story frames.

Because SDXL~\cite{sdxl} and FLUX~\cite{flux} use different text encoders, we adopt an encoder-specific integration strategy.

\paragraph{SDXL Integration.}

SDXL \cite{sdxl} contains two CLIP text encoders. We apply the Sidecar semantic extraction and injection independently to both encoders. Given the augmented token representations
$\widetilde{\mathbf{H}}^{(1)}_i$ and
$\widetilde{\mathbf{H}}^{(2)}_i$, the final text condition follows the original SDXL formulation:
\begin{equation}
    \widetilde{\mathbf{H}}^{\mathrm{SDXL}}_i
    =
    \operatorname{Concat}
    \left(
        \widetilde{\mathbf{H}}^{(1)}_i,
        \widetilde{\mathbf{H}}^{(2)}_i
    \right),
    \label{eq:sdxl_condition}
\end{equation}
where concatenation is performed along the feature dimension. The pooled text representation is obtained following the standard SDXL pipeline. Sidecar is applied only to the positive text condition, while the negative prompt is encoded without semantic augmentation.

\paragraph{FLUX Integration.}

FLUX~\cite{flux} uses a CLIP encoder and a T5 encoder to construct its text condition. We use different semantic augmentation strategies for the two encoders. For CLIP, we apply the layer-wise Sidecar mechanism described above.
For T5, we concatenate the first story prompt $P_0$ with the current prompt $P_i$ before encoding:
\begin{equation}
    \mathbf{Z}^{+}_i
    =
    \mathcal{E}_{\mathrm{T5}}
    \left(
        [P_0;P_i]
    \right),
    \label{eq:t5_joint_encoding}
\end{equation}
After joint contextual encoding, we retain only the hidden states corresponding to the current prompt:
\begin{equation}
    \widetilde{\mathbf{Z}}_i
    =
    \operatorname{Slice}_{P_i}
    \left(
        \mathbf{Z}^{+}_i
    \right).
    \label{eq:t5_slicing}
\end{equation}

Although the output corresponding to $P_0$ is removed, the tokens from the first prompt contribute to the T5 self-attention computation. The retained states of $P_i$ are therefore contextualized by both the initial character information and the current narrative. The resulting representation has the same sequence length as the standard T5 encoding of $P_i$, preserving compatibility with the original FLUX pipeline.

We use this encoder-specific strategy based on the empirical results. Sidecar enables layer-wise semantic interaction in the CLIP pathway, while T5 jointly contextualizes the initial and current prompts through its native self-attention. The longer context supported by T5 makes prompt concatenation practical, while avoiding direct modification of its intermediate hidden states. The resulting CLIP and T5 representations provide complementary semantic conditioning.

Overall, Sidecar introduces no trainable parameters and requires no additional optimization. It preserves the input-output interface of the original text-conditioning pipeline, allowing it to be integrated with both vanilla generation backbones and existing training-free storytelling methods.

\begin{figure*}
    \centering
    \includegraphics[width=\linewidth]{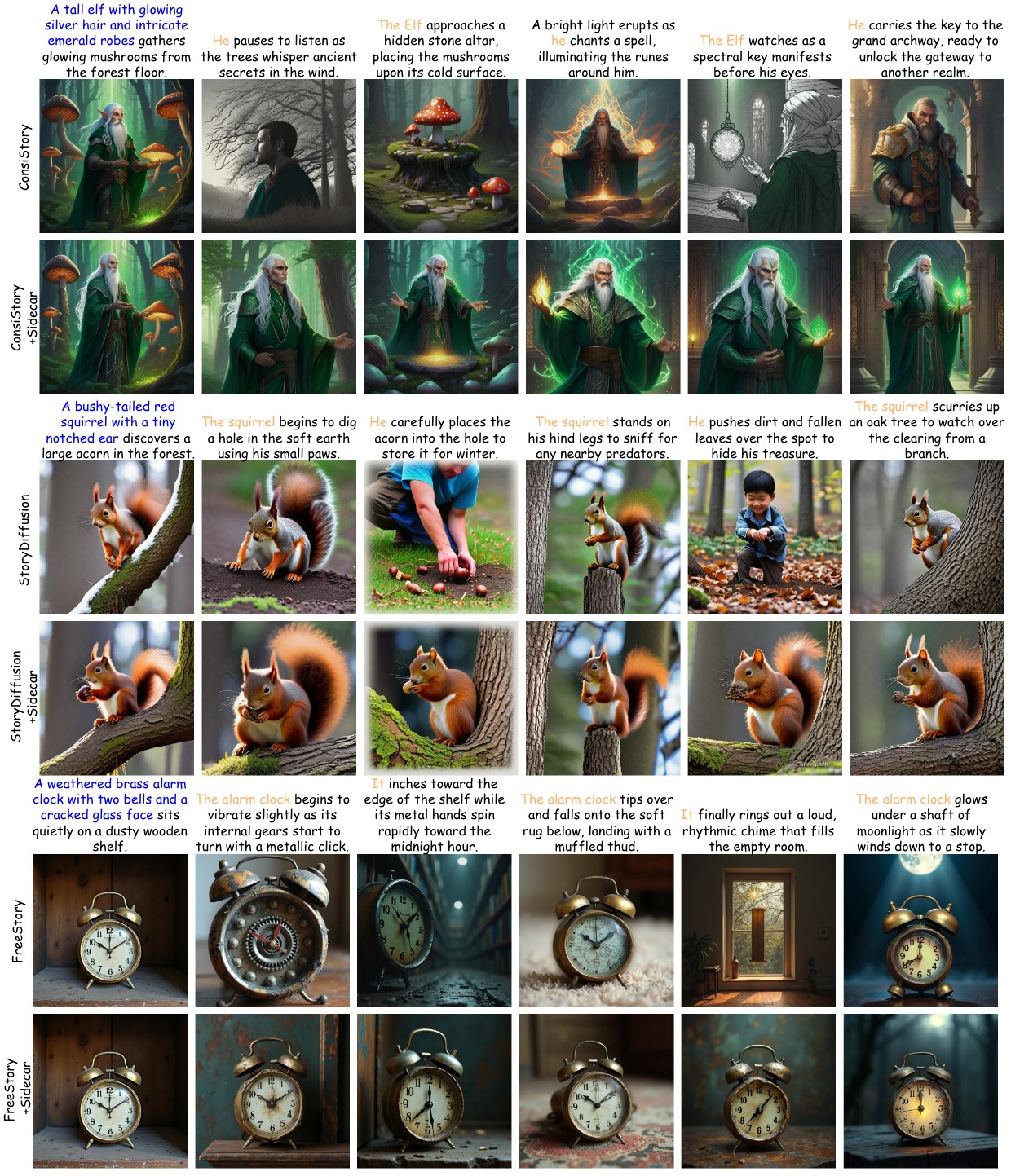}
    \caption{\textbf{Qualitative comparison on FreeStoryBench.} Each story provides a complete character description only in the first prompt, while later prompts use type or pronouns. Compared with the original baselines, Sidecar better preserves character consistency across different scenes. Sidecar also maintains text-image alignment, indicating that the improved consistency does not prevent the character or scene from changing according to the narrative. Additional qualitative results are provided in the supplementary material.}
    \label{fig:main}
\end{figure*}


\newcommand{\basescore}[1]{%
  \makebox[3.2em][r]{#1}%
  \makebox[4.2em][l]{}%
}

\newcommand{\sidecarscore}[2]{%
  \makebox[3.2em][r]{#1}%
  \makebox[4.2em][l]{\footnotesize~(#2)}%
}

\begin{table*}[t]
\centering
\small
\caption{
\textbf{Quantitative results on FreeStoryBench.}
We evaluate SDXL-based and FLUX-based methods with and without Sidecar. Higher CLIP and DINO scores are better, while lower DreamSim is better. For each Sidecar variant, the value in parentheses denotes the change relative to its corresponding baseline. Sidecar consistently improves character consistency and prompt-image alignment across all evaluated models, with particularly large gains on DINO and DreamSim.
}
\label{tab:main_results}

\begin{tabular}{llccc}
\toprule
& \textbf{Method}
& \textbf{CLIP} $\uparrow$
& \textbf{DINO} $\uparrow$
& \textbf{DreamSim} $\downarrow$ \\
\midrule

\multirow{7}{*}{\rotatebox{90}{ \small\shortstack{\scriptsize\textit{SDXL-based}\\models} }}
& \cellcolor{gray!6} OnePromptOneStory~\cite{onepromptonestory}
& \cellcolor{gray!6} \basescore{0.791}
& \cellcolor{gray!6} \basescore{0.361}
& \cellcolor{gray!6} \basescore{0.495} \\

\cline{2-5}

& \cellcolor{gray!6} SDXL
& \cellcolor{gray!6} \basescore{0.776}
& \cellcolor{gray!6} \basescore{0.308}
& \cellcolor{gray!6} \basescore{0.540} \\

& \cellcolor{gray!14} SDXL + Sidecar
& \cellcolor{gray!14} \sidecarscore{0.859}{+0.083}
& \cellcolor{gray!14} \sidecarscore{0.621}{+0.313}
& \cellcolor{gray!14} \sidecarscore{0.333}{-0.207} \\

& \cellcolor{gray!6} StoryDiffusion~\cite{storydiffusion}
& \cellcolor{gray!6} \basescore{0.803}
& \cellcolor{gray!6} \basescore{0.410}
& \cellcolor{gray!6} \basescore{0.483} \\

& \cellcolor{gray!14} StoryDiffusion + Sidecar
& \cellcolor{gray!14} \sidecarscore{\textbf{0.886}}{+0.083}
& \cellcolor{gray!14} \sidecarscore{\textbf{0.702}}{+0.292}
& \cellcolor{gray!14} \sidecarscore{0.272}{-0.211} \\

& \cellcolor{gray!6} ConsiStory~\cite{consistory}
& \cellcolor{gray!6} \basescore{0.779}
& \cellcolor{gray!6} \basescore{0.349}
& \cellcolor{gray!6} \basescore{0.505} \\

& \cellcolor{gray!14} ConsiStory + Sidecar
& \cellcolor{gray!14} \sidecarscore{0.880}{+0.101}
& \cellcolor{gray!14} \sidecarscore{0.692}{+0.343}
& \cellcolor{gray!14} \sidecarscore{\textbf{0.264}}{-0.241} \\

\midrule

\multirow{4}{*}{\rotatebox{90}{ \small\shortstack{\scriptsize\textit{FLUX-based}\\models} }}

& \cellcolor{gray!6} FLUX
& \cellcolor{gray!6} \basescore{0.776}
& \cellcolor{gray!6} \basescore{0.323}
& \cellcolor{gray!6} \basescore{0.523} \\

& \cellcolor{gray!14} FLUX + Sidecar
& \cellcolor{gray!14} \sidecarscore{0.842}{+0.066}
& \cellcolor{gray!14} \sidecarscore{0.575}{+0.252}
& \cellcolor{gray!14} \sidecarscore{0.346}{-0.177} \\

& \cellcolor{gray!6} FreeStory~\cite{freestory}
& \cellcolor{gray!6} \basescore{0.801}
& \cellcolor{gray!6} \basescore{0.421}
& \cellcolor{gray!6} \basescore{0.448} \\

& \cellcolor{gray!14} FreeStory + Sidecar
& \cellcolor{gray!14} \sidecarscore{0.869}{+0.068}
& \cellcolor{gray!14} \sidecarscore{0.649}{+0.228}
& \cellcolor{gray!14} \sidecarscore{0.291}{-0.157} \\

\bottomrule
\end{tabular}
\end{table*}

\section{Experiments}

\subsection{Experimental Setup}

\paragraph{Dataset.} We evaluate our method on the single-character subset of FreeStoryBench~\cite{freestory}, which contains 100 stories with six scenes per story. Each story can be rendered under six prompting settings that differ in how the character is referenced across scenes. We use the \emph{mixed} setting, where the complete character description appears only in the first prompt, while subsequent prompts refer to the character using a mixture of type-level mentions and pronouns. This setting reflects natural storytelling and provides a challenging testbed for character consistency because later prompts omit the explicit character semantic information.

\paragraph{Baselines.}
We compare Sidecar with both SDXL-based and FLUX-based generation models. For the SDXL-based models, we evaluate vanilla SDXL~\cite{sdxl}, StoryDiffusion~\cite{storydiffusion}, and ConsiStory~\cite{consistory}. For the FLUX-based models, we evaluate vanilla FLUX~\cite{flux} and FreeStory~\cite{freestory}. We additionally compare with OnePromptOneStory~\cite{onepromptonestory}. Since it concatenates all story prompts into a single input and directly modifies the text embeddings, its formulation differs from the targeted free-form setting and is not readily compatible with our proposed Sidecar. We therefore report it as an independent baseline without applying Sidecar. 

\paragraph{Metrics.}
We use three metrics to evaluate the generated story images. CLIP \cite{clip} similarity measures the semantic alignment between each generated image and its corresponding text prompt. DINO \cite{dino} similarity evaluates character consistency across frames within the same story. DreamSim \cite{dreamsim} measures perceptual similarity between generated character appearances, where a lower value indicates better consistency. To reduce the influence of background similarity, we use Grounded SAM~\cite{groundedsam} to segment the target character in each image, and replace the background region with random noise before computing DINO and DreamSim.

\paragraph{Implementation Details.}
Sidecar is applied at inference time without introducing trainable parameters or modifying pretrained image-generation models. It operates on all CLIP Transformer layers and only augments the positive text-conditioning path. For SDXL, Sidecar is applied to both CLIP encoders. For FLUX, it is applied to CLIP, while T5 jointly encodes $[P_0;P_i]$ with a maximum length of 512 tokens and retains only the hidden states of $P_i$. We use the default inference settings from the released baseline implementations and keep the generation settings fixed when comparing each baseline with its Sidecar variant. All experiments are conducted on a single NVIDIA L40S GPU. 
Our code will be made publicly available upon acceptance.

\subsection{Results on FreeStoryBench}

Table~\ref{tab:main_results} shows the quantitative results on FreeStoryBench. We report the performance of each baseline before and after adding Sidecar. The results show that Sidecar consistently improves all evaluated baselines across both SDXL-based and FLUX-based models. For the SDXL backbone, Sidecar improves vanilla SDXL from 0.776 to 0.859 in CLIP, from 0.308 to 0.621 in DINO, and reduces DreamSim from 0.540 to 0.333. Similar improvements are also observed when Sidecar is applied to StoryDiffusion and Consistory. 
For FLUX-based models, Sidecar also shows consistent improvement, indicating that Sidecar is not limited to a specific diffusion architecture, and can be applied to both UNet-based and transformer-based generation backbones.

The improvements are especially significant on the consistency metrics. 
This suggests that the proposed semantic sidecar effectively helps the model preserve character identity across different story frames. Moreover, Sidecar also improves CLIP similarity for every baseline. This shows that the improved consistency does not come from simply forcing visual similarity between frames, but also improves the semantic alignment between prompt and image.

We also measure the computational overhead introduced by Sidecar across different baseline methods. Sidecar introduces negligible computational overhead, as it only augments the text-encoding process while leaving the generation backbone unchanged. Across the evaluated baselines, inference time and peak GPU memory remain largely comparable after enabling Sidecar; for example, on SDXL, runtime increases by only $0.2\%$ with unchanged peak memory, while on ConsiStory the increases are $0.9\%$ and $0.5\%$, respectively. Full efficiency results are reported in Appendix.


Overall, these results show that Sidecar serves as a plug-and-play semantic augmentation module for free-form story generation. Across different baselines and backbones, Sidecar consistently improves character consistency with negligible additional computational cost, while maintaining and in all cases improving prompt-image alignment.

\subsection{Ablation Study}

\begin{table}[t]
\centering
\setlength{\tabcolsep}{3pt}
\caption{\textbf{Effect of Sidecar under different character-reference types.} $\Delta$ reports the improvement introduced by Sidecar. For DreamSim, improvement is measured by the reduction in distance.}
\label{tab:reference_type_ablation}
\resizebox{\columnwidth}{!}{
\setlength{\tabcolsep}{3pt}
\begin{tabular}{clccc|ccc}
\toprule
& & \multicolumn{3}{c|}{\textbf{Type-level Mention}}
& \multicolumn{3}{c}{\textbf{Pronoun}} \\
\cmidrule(lr){3-5}
\cmidrule(lr){6-8}
 & \textbf{Metric}
& \textbf{Baseline} & \textbf{+Sidecar} & \textbf{$\Delta$}
& \textbf{Baseline} & \textbf{+Sidecar} & \textbf{$\Delta$} \\
\midrule
\multirow{3}{*}{\rotatebox[origin=c]{90}{\footnotesize FreeStory}}
& CLIP $\uparrow$
& 0.844 & 0.850 & 0.007
& 0.771 & 0.849 & \textbf{0.078} \\
& DINO $\uparrow$
& 0.614 & 0.690 & 0.077
& 0.334 & 0.643 & \textbf{0.309} \\
& DreamSim $\downarrow$
& 0.331 & 0.270 & 0.061
& 0.506 & 0.301 & \textbf{0.206} \\
\midrule
\multirow{3}{*}{\rotatebox[origin=c]{90}{\footnotesize SDXL}}
& CLIP $\uparrow$
& 0.810 & 0.862 & 0.052
& 0.712 & 0.864 & \textbf{0.151} \\
& DINO $\uparrow$
& 0.503 & 0.636 & 0.133
& 0.201 & 0.602 & \textbf{0.401} \\
& DreamSim $\downarrow$
& 0.438 & 0.322 & 0.116
& 0.614 & 0.341 & \textbf{0.273} \\
\bottomrule
\end{tabular}
} 
\end{table}

\paragraph{Effect of Reference Type.} Table~\ref{tab:reference_type_ablation} analyzes the effect of Sidecar under type-level mentions and pronouns. We report results for SDXL and FreeStory to cover two different generation backbones while also including both a vanilla T2I model and a dedicated storytelling method. This provides a representative evaluation of whether the observed behavior generalizes across model families and generation settings. Sidecar improves all metrics for both FreeStory and SDXL, with consistently larger gains for pronoun-based prompts. On FreeStory, the DINO improvement increases from 0.077 for type-level mentions to 0.309 for pronouns, while the DreamSim improvement increases from 0.061 to 0.206. A similar trend is observed on SDXL. These results support our hypothesis that Sidecar is particularly beneficial when the prompt contains less explicit identity information.

\paragraph{Text-Encoder Integration Strategies.}
Table~\ref{tab:encoder_integration_ablation} compares different strategies for incorporating the character semantics into the CLIP and T5 encoders of FreeStory. Applying Sidecar only to CLIP improves all metrics over the original FreeStory, showing that layer-wise semantic augmentation is effective for CLIP. T5 concatenation alone yields only modest improvements, while directly applying Sidecar to T5 degrades performance, suggesting that the same injection strategy does not transfer uniformly across text encoders.
Joint prompt concatenation for both encoders provides a strong alternative. However, with T5 concatenation fixed, replacing CLIP concatenation with Sidecar further improves all metrics from $0.860/0.594/0.324$ to $0.869/0.649/0.291$ for CLIP/DINO/DreamSim. This controlled comparison isolates the contribution of CLIP Sidecar and shows that the gains of the FLUX-based integration cannot be attributed to T5 concatenation alone.

The results suggest that CLIP and T5 benefit from different integration strategies. For CLIP, direct concatenation is constrained by its 77-token limit and may truncate scene information, especially for multi-character prompts. Sidecar avoids consuming this token budget by separately encoding the reference and injecting only character-description representations. In contrast, T5 supports a much longer context, making joint prompt concatenation more practical. Our ablation also shows that Sidecar is less effective than concatenation for T5. The best performance is therefore achieved by combining CLIP Sidecar with T5 concatenation.

\begin{table}[t]
\centering
\caption{
\textbf{Ablation of text-encoder integration strategies on FreeStory.}
The gray row denotes the FreeStory setting, while the green row denotes the FreeStory+Sidecar setting.
}
\label{tab:encoder_integration_ablation}
\resizebox{0.96\columnwidth}{!}{
\setlength{\tabcolsep}{4pt}
\begin{tabular}{cc|ccc}
\toprule
\textbf{CLIP} & \textbf{T5}
& \textbf{CLIP} $\uparrow$ & \textbf{DINO} $\uparrow$ & \textbf{DreamSim} $\downarrow$ \\
\hline 
\rowcolor{gray!12} Original & Original
& 0.801 & 0.421 & 0.448 \\
Sidecar & Original
& 0.830 & 0.481 & 0.413 \\
Original & Concatenation
& 0.809 & 0.432 & 0.444 \\
Original & Sidecar
& 0.750 & 0.285 & 0.548 \\
Sidecar & Sidecar
& 0.794 & 0.514 & 0.399 \\
Concatenation & Concatenation
& 0.860 & 0.594 & 0.324 \\
\rowcolor{green!12} Sidecar & Concatenation
& 0.869 & 0.649 & 0.291 \\
\bottomrule
\end{tabular}
}
\end{table}

\begin{table}[t]
\centering
\caption{
\textbf{Effect of semantic interaction location on SDXL.} Post-encoder replacement substitutes the final mention representations with those of the full description, whereas Sidecar enables layer-wise interaction between the description and current prompt. Both are applied to both CLIP encoders. 
}
\label{tab:semantic_interaction_location}
\resizebox{0.96\columnwidth}{!}{
\setlength{\tabcolsep}{4pt}
\begin{tabular}{l|ccc}
\toprule
\textbf{Method}
& \textbf{CLIP} $\uparrow$ & \textbf{DINO} $\uparrow$ & \textbf{DreamSim} $\downarrow$ \\
\hline 
\rowcolor{gray!12}
SDXL
& 0.776 & 0.308 & 0.540 \\
SDXL + Post-Encoder Expansion
& 0.835 & 0.488 & 0.407 \\
\rowcolor{green!12}
SDXL + Sidecar
& 0.859 & 0.621 & 0.333 \\
\bottomrule
\end{tabular}
}\vspace{-0.8em}
\end{table}

\paragraph{Effect of Semantic Interaction Location.}
Table~\ref{tab:semantic_interaction_location} compares Sidecar with a post-encoder semantic expansion baseline on SDXL. Specifically, it independently encodes the initial description and current prompt, then replaces the final CLIP representations of the current character mention with the final representations of the complete description. Although this strategy restores the omitted semantics and bypasses CLIP's input-length constraint, the description and current-prompt representations do not interact through text-encoder self-attention.

Post-encoder expansion improves over the original SDXL baseline, showing that directly restoring the missing information is beneficial. Sidecar achieves stronger overall performance, suggesting that semantic availability alone is insufficient. Allowing the reused description representations to interact with the current action and scene within each CLIP layer provides more effective character conditioning than directly modifying the final text representation.

\subsection{Discussion}

Sidecar complements visual consistency methods by addressing missing identity semantics in later free-form prompts. Its gains on both vanilla generators and storytelling baselines suggest that text-side semantic reuse remains useful even when visual features are already propagated across frames. Larger improvements for pronouns support this interpretation, while the comparison with concatenation and post-encoder replacement indicates that layer-wise interaction within CLIP is more effective than simply providing the initial description. The different behaviors of CLIP and T5 further motivate encoder-specific integration. The improvements in CLIP similarity suggest that consistency is not achieved at the expense of overall prompt-image alignment.

\paragraph{Limitations.}
Sidecar relies on the initial prompt to provide a clear and accurate character description; ambiguous or incomplete descriptions may therefore limit its effectiveness. Our current evaluation focuses on stories with a single main character, and extending the method to multiple interacting characters would require reliable entity-specific extraction and disambiguation. In addition, CLIP retains its 77-token context limit, so very long initial prompts may truncate relevant character information. Our experiments demonstrate that Sidecar is particularly effective with CLIP-style text encoders, while extending its benefits to other encoder architectures remains a direction for future work.

\section{Conclusion} 
We introduced Sidecar, a training-free semantic augmentation module for free-form visual storytelling. Sidecar extracts identity-related representations from the initial character description and injects them into the encoding of later prompts, restoring information omitted by the reference mentions. Sidecar requires no additional training, introduces negligible computational overhead, and can be integrated with both vanilla diffusion models and existing storytelling methods. Experiments on FreeStoryBench show that Sidecar improves character consistency and prompt-image alignment across SDXL- and FLUX-based models. These results demonstrate that incomplete text semantics are an important source of identity drift and that text-side semantic augmentation provides a simple and complementary solution for more consistent story generation.
\newpage
{
    \small
    \bibliographystyle{ieeenat_fullname}
    \bibliography{main}
}

\clearpage
\appendix

\section{Multi-Character Evaluation}

We evaluate Sidecar on the multi-character subset of FreeStoryBench to examine whether entity-specific semantic augmentation remains effective when multiple characters appear in the same story. The subset contains 100 stories and 600 frames. For each character, we use its corresponding structured description span provided by FreeStoryBench and construct the Sidecar representation independently, allowing each entity to retrieve its own reference semantics without sharing description tokens.

Table~\ref{tab:multi_character} shows that Sidecar consistently improves performance across all evaluated baselines. DINO increases substantially for all compared baselines, with corresponding reductions in DreamSim. Importantly, CLIP alignment improves for every baseline, indicating that the gains in multi-character consistency do not come at the expense of overall prompt fidelity. These results demonstrate that Sidecar extends effectively to multi-character stories and can preserve entity-specific semantics without evident interference between characters.

\begin{table}[t]
\centering
\caption{Multi-character evaluation on FreeStoryBench. Higher is better for CLIP and DINO, while lower is better for DreamSim.}
\label{tab:multi_character}
\resizebox{\columnwidth}{!}{
\setlength{\tabcolsep}{0pt}
\begin{tabular}{llll}
\toprule
Method & CLIP $\uparrow$ & DINO $\uparrow$ & DreamSim $\downarrow$ \\
\midrule
\cellcolor{gray!6} SDXL
& \cellcolor{gray!6} 0.808
& \cellcolor{gray!6} 0.327
& \cellcolor{gray!6} 0.496 \\
\cellcolor{gray!14} SDXL + Sidecar
& \cellcolor{gray!14} 0.888 {\scriptsize $(+0.080)$}
& \cellcolor{gray!14} 0.506 {\scriptsize $(+0.179)$}
& \cellcolor{gray!14} 0.380 {\scriptsize $(-0.116)$} \\
\midrule
\cellcolor{gray!6} StoryDiffusion
& \cellcolor{gray!6} 0.813
& \cellcolor{gray!6} 0.388
& \cellcolor{gray!6} 0.477 \\
\cellcolor{gray!14} StoryDiffusion + Sidecar
& \cellcolor{gray!14} 0.884 {\scriptsize $(+0.071)$}
& \cellcolor{gray!14} 0.570 {\scriptsize $(+0.182)$}
& \cellcolor{gray!14} 0.340 {\scriptsize $(-0.137)$} \\
\midrule
\cellcolor{gray!6} ConsiStory
& \cellcolor{gray!6} 0.823
& \cellcolor{gray!6} 0.359
& \cellcolor{gray!6} 0.478 \\
\cellcolor{gray!14} ConsiStory + Sidecar
& \cellcolor{gray!14} 0.893 {\scriptsize $(+0.070)$}
& \cellcolor{gray!14} 0.589 {\scriptsize $(+0.230)$}
& \cellcolor{gray!14} 0.307 {\scriptsize $(-0.171)$} \\
\midrule
\cellcolor{gray!6} FLUX
& \cellcolor{gray!6} 0.799
& \cellcolor{gray!6} 0.360
& \cellcolor{gray!6} 0.471 \\
\cellcolor{gray!14} FLUX + Sidecar
& \cellcolor{gray!14} 0.850 {\scriptsize $(+0.051)$}
& \cellcolor{gray!14} 0.485 {\scriptsize $(+0.125)$}
& \cellcolor{gray!14} 0.384 {\scriptsize $(-0.087)$} \\
\midrule
\cellcolor{gray!6} FreeStory
& \cellcolor{gray!6} 0.845
& \cellcolor{gray!6} 0.463
& \cellcolor{gray!6} 0.392 \\
\cellcolor{gray!14} FreeStory + Sidecar
& \cellcolor{gray!14} 0.867 {\scriptsize $(+0.022)$}
& \cellcolor{gray!14} 0.522 {\scriptsize $(+0.059)$}
& \cellcolor{gray!14} 0.354 {\scriptsize $(-0.038)$} \\
\bottomrule
\end{tabular}}
\end{table}

\section{Effect of Sidecar Placement Across Encoder Layers}

We further study how the placement of Sidecar within the text encoder affects its performance. Using SDXL, we divide the CLIP text encoder into three equal depth ranges and apply Sidecar only to the first (Early), middle (Middle), or last (Late) one-third of the encoder layers. We compare these variants with the original SDXL baseline and our default configuration that applies Sidecar at all encoder layers. 

\begin{table}[t]
\centering
\caption{Ablation on Sidecar placement across CLIP encoder layers on SDXL. We only apply Sidecar to the first, middle, and last one-third of encoder layers, respectively.}
\label{tab:layer_ablation}
\resizebox{\columnwidth}{!}{
\setlength{\tabcolsep}{0pt}
\begin{tabular}{lccc}
\toprule
Method & CLIP-T $\uparrow$ & DINO $\uparrow$ & DreamSim $\downarrow$ \\
\midrule
\cellcolor{gray!6} SDXL
& \cellcolor{gray!6} \basescore{0.776}
& \cellcolor{gray!6} \basescore{0.308}
& \cellcolor{gray!6} \basescore{0.540} \\
 + Sidecar (Early 1/3)
& \sidecarscore{0.759}{-0.017}
& \sidecarscore{0.326}{+0.018}
& \sidecarscore{0.526}{-0.014} \\
 + Sidecar (Middle 1/3)
& \sidecarscore{0.785}{+0.009}
& \sidecarscore{0.401}{+0.093}
& \sidecarscore{0.500}{-0.040} \\
 + Sidecar (Late 1/3)
& \sidecarscore{0.840}{+0.064}
& \sidecarscore{0.544}{+0.236}
& \sidecarscore{0.394}{-0.146} \\
\cellcolor{gray!14} + Sidecar (All)
& \cellcolor{gray!14} \sidecarscore{0.859}{+0.083}
& \cellcolor{gray!14} \sidecarscore{0.621}{+0.313}
& \cellcolor{gray!14} \sidecarscore{0.333}{-0.207} \\
\bottomrule
\end{tabular}}
\end{table}

As shown in Table~\ref{tab:layer_ablation}, the effectiveness of Sidecar increases substantially with encoder depth. Applying Sidecar only to the early layers provides little improvement over the baseline, whereas the middle layers yield moderate gains and the late layers produce a much larger improvement in both character consistency and prompt alignment. This trend suggests that the higher-level contextual representations formed in deeper CLIP layers are particularly important for recovering character semantics. Nevertheless, applying Sidecar throughout the entire encoder consistently achieves the best performance.

\begin{table}[t]
\centering
\caption{Computational efficiency on the full dataset evaluation. Runtime is averaged over all processed samples, and GPU memory reports the peak allocated memory during inference.}
\label{tab:efficiency_full}
\resizebox{\columnwidth}{!}{
\setlength{\tabcolsep}{2pt}
\begin{tabular}{lccrr}
\toprule
Method & Peak Mem. (GB) & Avg. Time (s) & $\Delta$Mem. & $\Delta$Time \\
\midrule
SDXL & 10.49 & 35.68 & -- & -- \\
SDXL + Sidecar & 10.49 & 35.76 & 0.0\% & +0.2\% \\
\midrule
StoryDiffusion & 36.67 & 61.13 & -- & -- \\
StoryDiffusion + Sidecar & 36.68 & 62.61 & 0.0\% & +2.4\% \\
\midrule
ConsiStory & 21.66 & 151.41 & -- & -- \\
ConsiStory + Sidecar & 21.77 & 152.72 & +0.5\% & +0.9\% \\
\midrule
FLUX & 33.83 & 158.18 & -- & -- \\
FLUX + Sidecar & 34.05 & 158.71 & +0.7\% & +0.3\% \\
\midrule
FreeStory & 35.07 & 209.20 & -- & -- \\
FreeStory + Sidecar & 35.07 & 214.74 & 0.0\% & +2.6\% \\
\bottomrule
\end{tabular}}
\end{table}

\section{Computational Efficiency}

We further evaluate the computational cost of Sidecar by measuring peak allocated GPU memory and average inference time under the same generation settings. Table~\ref{tab:efficiency_full} reports the results for each baseline and its Sidecar-augmented counterpart. Overall, Sidecar introduces only marginal computational overhead across all evaluated methods.
Peak GPU memory remains nearly unchanged after enabling Sidecar: SDXL, StoryDiffusion and FreeStory show no measurable increase, while ConsiStory and FLUX increase by only $0.5\%$, and $0.7\%$, respectively. The additional inference time is similarly small, ranging from $0.2\%$ on SDXL to at most $2.6\%$ on FreeStory. These results are consistent across both vanilla generators and existing consistency methods.

Overall, Sidecar is computationally lightweight, adding minimal memory and runtime overhead while substantially improving character consistency. This efficiency follows from its design, which augments only the text-conditioning process while leaving the underlying image-generation backbone unchanged.

\begin{figure*}[t]
    \centering
    \includegraphics[width=\linewidth]{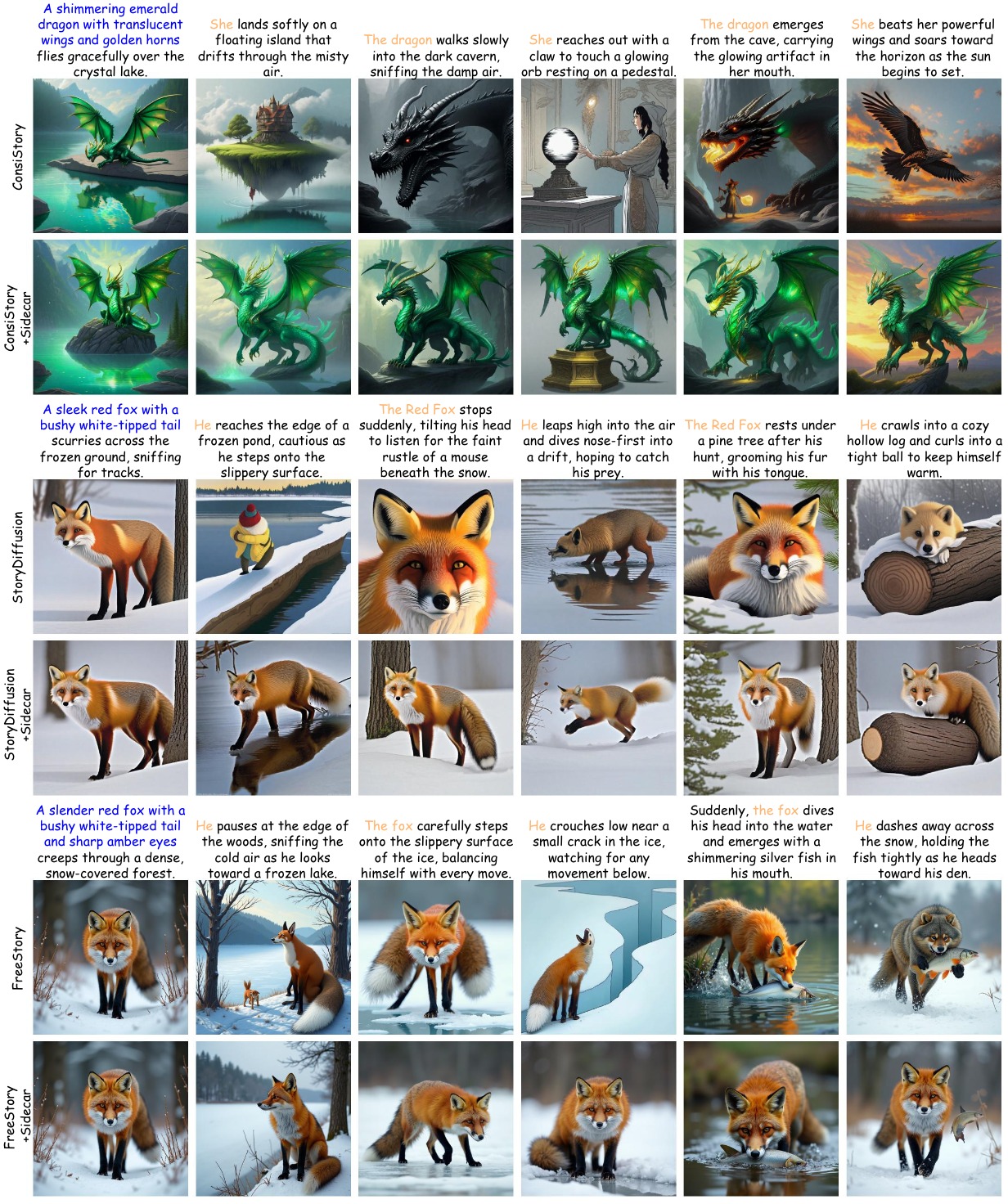}
    \caption{\textbf{Qualitative comparison} on FreeStoryBench. }
    \label{fig:extra}
\end{figure*}
\begin{figure*}[t]
    \centering
    \includegraphics[width=\linewidth]{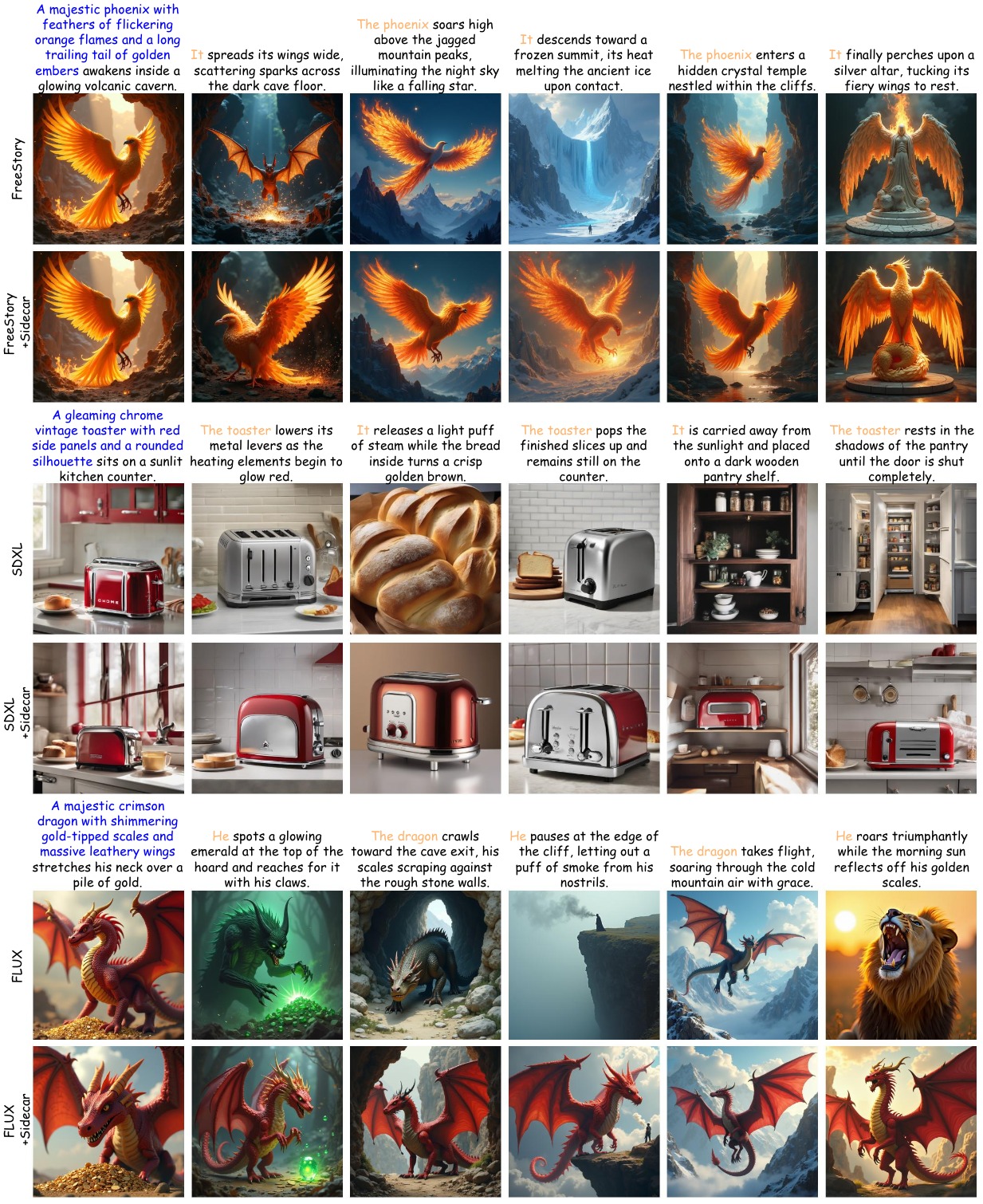}
    \caption{\textbf{Qualitative comparison} on FreeStoryBench. Sidecar recovers identity semantics omitted from later prompts. Since vanilla T2I models do not contain an explicit cross-frame consistency mechanism, Sidecar improves identity preservation but does not guarantee complete visual consistency.}
    \label{fig:extra}
\end{figure*}
\begin{figure*}[t]
    \centering
    \includegraphics[width=\linewidth]{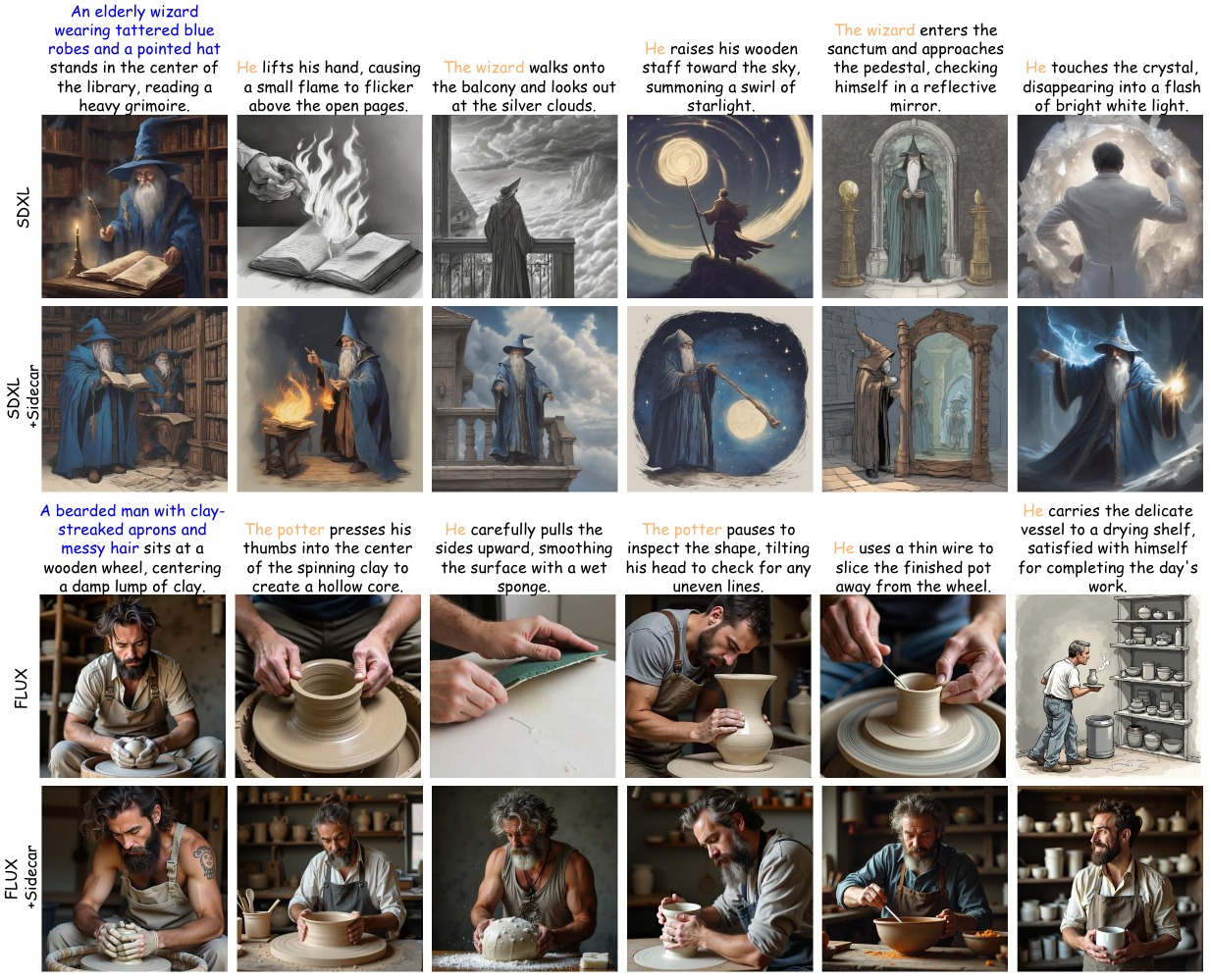}
    \caption{\textbf{Qualitative comparison} on FreeStoryBench. Sidecar recovers identity semantics omitted from later prompts. Since vanilla T2I models do not contain an explicit cross-frame consistency mechanism, Sidecar improves identity preservation but does not guarantee complete visual consistency.}
    \label{fig:extra}
\end{figure*}

\end{document}